\documentclass{article} % For LaTeX2e
\usepackage{iclr2022_conference_arxiv,times}

\usepackage{amsmath,amsfonts,bm}

\def\eqref#1{equation~\ref{#1}}
\def\1{\bm{1}}

\DeclareMathAlphabet{\mathsfit}{\encodingdefault}{\sfdefault}{m}{sl}
\SetMathAlphabet{\mathsfit}{bold}{\encodingdefault}{\sfdefault}{bx}{n}

\usepackage{hyperref}
\usepackage{url}
\usepackage{graphicx}
\usepackage{booktabs}
\usepackage{wrapfig}
\usepackage{lipsum}

\title{Constrained Mean Shift for Representation Learning}

\author{\fontsize{11}{11} \selectfont Ajinkya Tejankar$^{1,}\footnotemark[1]$ $\quad$
 Soroush Abbasi Koohpayegani$^{1,}\thanks{Equal contribution. Contact \{at6,soroush\}@umbc.edu}$ $\quad$ Hamed Pirsiavash$^{1,2}$   \\
\fontsize{11}{11} \selectfont $^{1}$ University of Maryland, Baltimore County $\quad$ $^{2}$University of California, Davis\\
}

\iclrfinalcopy % Uncomment for camera-ready version, but NOT for submission.
\begin{document}

\maketitle

\begin{abstract}
We are interested in representation learning from labeled or unlabeled data. Inspired by recent success of self-supervised learning (SSL), we develop a non-contrastive representation learning method that can exploit additional knowledge. This additional knowledge may come from annotated labels in the supervised setting or an SSL model from another modality in the SSL setting. Our main idea is to generalize the mean-shift algorithm by constraining the search space of nearest neighbors, resulting in semantically purer representations.
Our method simply pulls the embedding of an instance closer to its nearest neighbors in a search space that is constrained using the additional knowledge. By leveraging this non-contrastive loss, we show that the supervised ImageNet-1k pretraining with our method results in better transfer performance as compared to the baselines. Further, we demonstrate that our method is relatively robust to label noise. Finally, we show that it is possible to use the noisy constraint across modalities to train self-supervised video models.
\end{abstract}

\section{Introduction}
It is a common practice in visual recognition to pretrain a model using cross-entropy loss on some annotated data set (e.g., ImageNet) and then use the learned representations on a downstream task with limited annotated data. We are interested in improving this process by generalizing a recent self-supervised learning (SSL) idea to use other sources of knowledge, e.g., annotation.

Recently, we have seen great progress in self-supervised learning (SSL) methods that learn rich representations from unlabeled data. Such methods are important since they do not rely on manual annotation of data which can be costly, biased, or ambiguous. Hence, SSL representations may perform better than supervised ones in transferring to downstream visual recognition tasks.

Some popular recent SSL methods called ``contrastive'' assume that in the embedding space, an image should be closer to its own augmentation compared to some other random images \cite{he2020momentum}. Some other methods learn representations by clustering images together \citep{caron2018deep} using k-means-like algorithms that contrast between different clusters of images. 
This contrastive setting has been used in the deep learning community for a long time. Even the standard supervised learning with cross entropy loss maximizes the probability of the correct output while suppressing the probability of the wrong outputs automatically through the normalization in the SoftMax function. This can be seen as a form of contrast between correct and wrong categorization.

A recent SSL method, BYOL \citep{grill2020bootstrap}, showed that contrasting against random images is not really necessary and can result in a better performance. Only the augmentations of a single image are pulled closer. A very recent work, MSF \citep{koohpayegani2021mean}, has shown that this non-contrastive framework, e.g. BYOL, can be generalized by pulling an image to be closer to not only its augmentations but also its nearest neighbors. This can be seen as a mean-shift algorithm that groups similar images together, but is different from k-means based algorithms as it does not contrast against other groups of images or does not cluster images explicitly.

Our main idea is to design a non-contrastive representation learning method that can utilize other sources of knowledge, e.g., labels. We generalize the MSF method further by constraining the nearest neighbor (NN) search using the additional knowledge. This constraint can help in learning a less noisy grouping of images. For instance, in the supervised setting, when we search for the NNs of the query image to average them, we limit the search space to only the images that share the same label as the query image. Such a simple change makes sure that the NNs are all from the correct semantic category so that we do not pull the query towards images from other categories. 

Note that we group only a few NNs ($k$ in our method) from the query's category together instead of grouping all images together as done in standard cross entropy based learning. We believe this relaxes the learning by not forcing all images of a category to form a cluster or be on the same side of a hyper-plane. Such a relaxation can improve the model when using less robust sources of knowledge, e.g., noisy labels.

%(1) preserve the geometric structure of the category that may be important for fine-grained downstream tasks and (2) 

Moreover, our method is more general and can use other sources of knowledge for constraining the NN search. For instance, when training SSL models from videos, we use an already trained SSL method on one modality, e.g., RGB, to constrain the NN search on SSL training in a different modality, e.g., Flow.

Our method achieves superior results compared to the baselines in supervised setting with clean and noisy labels as well as self-supervised setting in video. 

%The source of knowledge that provides this constraint can be the categorical labels in the supervised setting or NN search using a SSL in a different modality in the self-supervised setting. 

%Our method uses the extra knowledge to limit the NN search space only rather than enforcing the model to predict that knowledge. An example is cross-entropy (Xent) based supervised setting. We believe our method can utilize less robust sources of knowledge, e.g., noisy annotation, weak annotation or even nearest neighbors in another modality. To this end, we show that our method outperforms the standard cross-entropy supervised model with a large margin when the annotation is noisy. Moreover, we show that in learning video representations, we can can use an already trained SSL model in one modality, e.g., RGB, to constrain the SSL learning in another modality, e.g., flow.

\begin{figure*}[t]
\centering
\includegraphics[width=0.99\linewidth]{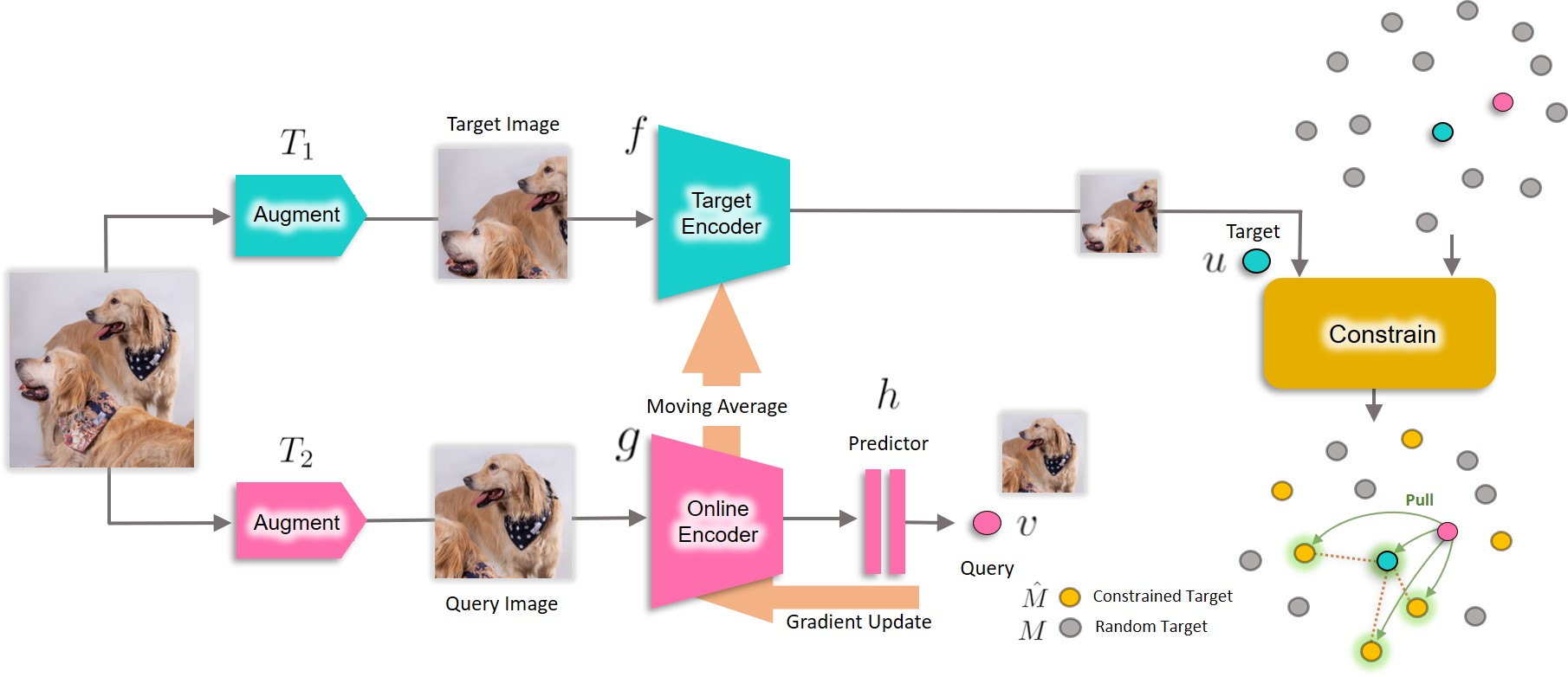}
\caption{{\bf Our method (CMSF):} We augment an image twice and pass through online encoder and target encoder followed by $\ell_2$ normalization to get $u$ and $v$. We want $v$ to be close to not only $u$, but also the nearest neighbors (NN) of $u$ to perform mean-shift. We constrain the NN pool using some extra knowledge, e.g., image labels, to improve the purity of the NNs. In supervised setting, we use images from the same category only (yellow points) to do NN search. We show that the constrain can come from noisy labels or NN search using a pre-trained SSL embedding on another modality.}
\label{fig:teaser}
\vspace{-.1in}
\end{figure*}

\section{Method}
%Following MSF \citep{} and MoCo \citep{}, we assume two embedding networks: a target encoder $f(.)$ with parameters $\theta_f$ and an online encoder $g(.)$ with parameters $\thata_g$. The online encoder is directly updated using backpropagation while the target encoder is updated as a slowly moving average of the online encoder: $\theta_f << m\theta_f + (1-m)\thata_g. This is the momentum idea introduced in \citep{MoCo}.

Similar to MSF \citep{koohpayegani2021mean}, given a query image, we are interested in pulling its embedding closer to the mean of the embeddings of its nearest neighbors (NNs). However, unlike MSF, we assume there is another source of knowledge that can constrain the set of data points in which we search for the NNs. This constraint can come from labels in the supervised setting or NNs on another modality in self-supervised learning for multi-modal data, e.g., videos. %In the regular MSF algorithm \citep{koohpayegani2021mean}, the set of NNs might not be all semantically related to each other. In that case, pulling the embedding of the query towards the average of those NNs can introduce some noise that may hurt the quality of the learned representations. 
%The hope is that by using our constraint we can improve the purity of the set of NNs where the purity is defined as the percentage of the NNs being from the same semantic category as the query image. For instance, in the supervised setting, the purity will be always 100\% since the constraint forces the NNs to share the same label as the query image.

To increase the size of the nearest neighbor pool, inspired by \citet{he2020momentum}, we use a large queue filled with the most recent training images as the memory bank. We maintain a slowly evolving average of the embedding model similar to \citet{he2020momentum}. Note that the computational costs of finding NNs is negligible compared to the overall training cost \citep{koohpayegani2021mean}, and they are needed in any method that uses a memory bank, e.g., MoCo \citep{he2020momentum}. %Since the embedding network is evolving over time, we maintain a slowly evolving average of the embedding model and use this slow encoder in finding the nearest neighbors.

Hence, we assume two embedding networks: a target encoder $f(.)$ with parameters $\theta_f$ and an online encoder $g(.)$ with parameters $\theta_g$. The online encoder is directly updated using backpropagation while the target encoder is updated as a slowly moving average of the online encoder: $\theta_f \leftarrow m\theta_f + (1-m)\theta_g$ where $m$ is close to $1$. This is the momentum idea introduced in \citet{he2020momentum}. We add a predictor head $h(.)$ \citep{grill2020bootstrap} to the end of the online encoder so that pulling the embeddings together encourages one embedding to be predicted by the other one and not necessarily encouraging the two embeddings to be equal. In the experiments, we use a two-layer MLP for $h(.)$.

\begin{figure*}[t]
\centering
\includegraphics[width=.9\linewidth]{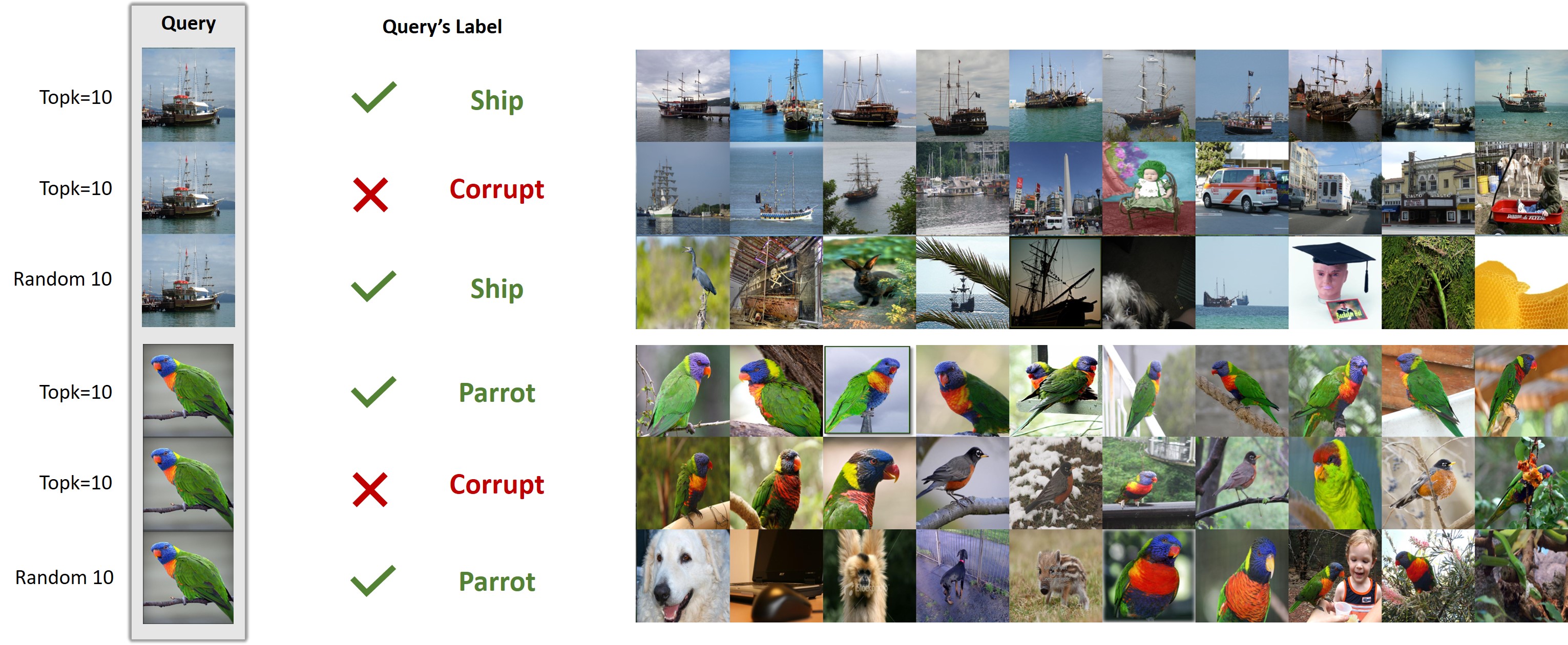}
\caption{{\bf top-$10$ vs. random 10 on ImageNet100 with 50\% noisy labels:} We show NN search results (sorted from left to right) for a query on the set of images that share the same label with the query. When the query is correctly labeled (Row 1), most NN results are from the same category which is good. When the query is not correctly labeled (Row 2), the few top results are still from the actual category of the query which is good. However, on Row 3, almost half of random results are not from the correct category. This is not surprising, but shows why CMSF with top-$k$ (Rows 1 and 2) performs better than CMSF with top-$all$ and standard cross-entropy supervised learning (Row 3).}
\label{fig:visualize_nns}
\vspace{-.1in}
\end{figure*}

{\bf Supervised-setting:} Given a query image $x$, we augment it twice with $T_1(.)$ and $T_2(.)$, feed them to the encoders, and normalize them with their $\ell_2$ norm to get $u=\frac{f(T_1(x))}{||f(T_1(x))||_2}$ and $v=\frac{h(g(T2(x)))}{||h(g(T2(x)))||_2}$. We add $u$ to the memory bank $M$ and remove the oldest entries to maintain a limited size for $M$. Since in the supervised setting, we know the label for each image, we choose the subset of $M$ that shares the same label with the query image to get $\hat M$. Then, we find top-$k$ neighbors of $u$ in $\hat M$ including $u$ itself and call it $S=\{z_i\}_{i=1}^k$. Finally, we update $g(.)$ by minimizing the following loss and update $f(.)$ with the momentum update. Note that self-supervised mean-shift algorithm in \citet{koohpayegani2021mean} is a specific case of our algorithm in which the constraint does not limit the nearest neighbor search, i,e. $\hat M=M$. 
\vspace{-.25in}

\begin{equation}
\label{eqn:eq1}
L=\frac{1}{k}\sum_{i=1}^k v^Tz_i
\end{equation}
\vspace{-.1in}

% $$L=\frac{1}{K}\sum_{i=1}^K v^Tz_i$$

In top-$all$ variation of our method, $k$ is equal to the total size of $\hat M$. Note that since $u$ itself is included in the nearest neighbor search, by limiting the size of the constrained set $\hat M$ to one (top-$1$), the method will be identical to BYOL \citep{grill2020bootstrap} and by setting $\hat M=M$, it will be identical to self-supervised mean-shift \citep{koohpayegani2021mean}. Hence, our method covers a larger spectrum by defining the constrained set. Unlike cross-entropy, our method does not encourage collapsing the whole class into one cluster. If the category has a multi-modal distribution, our method may group samples of each mode together without necessarily mixing the modes.

{\bf Supervised constraint with noisy labels:} Our method in the supervised setting uses the labels to constrain the mean-shift algorithm only rather than enforcing the labels directly in the loss function as done in standard cross-entropy learning. Hence, the constraint does not need to be strictly aligned with the semantic categories. Therefore, our method can benefit even from some weak signal provided by the constraint. For instance, we can use noisy labels to provide the constraint. We do extensive experiments with this setting and show that NNs are key to being robust against noise. Figure \ref{fig:visualize_nns} shows example NN search results for 50\% noisy labeled dataset.

{\bf Cross-modal constraint for self-supervised learning:} The constraint in our method is not limited to categorical semantic labels as discussed above. We can use nearest neighbor search in another modality to provide the constraint. For instance, in learning rich representations from unlabeled videos, we train a model for the RGB input and use it in a frozen form to constrain the memory bank in training a model for the flow input.

Assuming that $c(.)$ is an already trained embedding for RGB input, we want to train target encoder $f(.)$ and online encoder $g(.)$ for the flow input. Given a query video with RGB component $x_c$ and flow component $x$, we calculate $u$ and $v$ similar to the supervised setting using flow models $f(.)$ and $g(.)$ on the flow input $x$, and maintain the flow memory bank $M$ with target embeddings $u$.
% followed by augmentation, prediction MLP, and normalization. 

Since we do not have labels, we cannot simply construct $\hat M$ from $M$ as done in the supervised setting, so we feed the RGB component $x_c$ of the video to the frozen $c(.)$ encoder with the same augmentation and normalization to get the embedding $u_c$, and maintain a memory bank $M_c$ in which the data points follow the same ordering as in $M$. Then, we find $n$ nearest neighbors of $u_c$ in $M_c$ and use their indices to construct $\hat M$ from $M$. Finally, we use $\hat M$ to minimize the same loss as in Eq \ref{eqn:eq1} by finding top-$k$ nearest neighbors of $u$ in $\hat M$. 

Note that we calculate the flow using an unsupervised optical flow algorithm from RGB frames, so RGB and flow are not two distinct modalities as the flow can be calculated from the RGB modality. However, our method does not use this dependency and can be used for two distinct modalities.

\subsection{Supervised Setting}
\label{sec:supervised}

\subsubsection{Baselines}
{\bf Cross entropy (Xent):}  Xent \citep{NIPS1987_eccbc87e,levin1988accelerated,rumelhart1986learning} is a popular method for training standard supervised models.

\textbf{Supervised Contrastive (SupCon): } This method extends the instance discrimination framework from self-supervised learning to supervised learning \citep{khosla2020supervised}. It is a contrastive setting in which the positive set contains all images from the same category. The top-$all$ variation of our method is similar to SupCon without any contrast.

\textbf{Prototypical Networks (ProtoNW): } In order to further study the effect of contrast, we design another contrastive version of our top-$all$ variation. We calculate a prototype for each class by averaging all its instances in the memory bank. Then, similar to prototypical networks \citep{snell2017prototypical}, we compare the input with all prototypes by passing their temperature-scaled cosine distance through a SoftMax layer to get probabilities. Finally, we minimize the cross-entropy loss. Note that this method is still contrastive in nature because of the SoftMax operation.

\textbf{Frozen Prototype (FrzProto): } We randomly initialize a set of prototype embeddings for each class and freeze them throughout training. The prototypes are used as targets for regressing the output embeddings from the backbone network using Cosine similarity loss. This method is very similar to the noise-as-target method \citep{bojanowski2017unsupervised}. It can be thought of as the above ProtoNW method but with random class prototypes. Note that since the prototypes are frozen and initialized randomly, even semantically related categories will be far away from each other. Thus, pulling an embedding close to its target class embedding implies that the embedding is pushed away from other class embeddings. This makes the method contrastive. Surprisingly, even frozen prototypes work remarkably well when we change the final linear FC layer to a 2-layer MLP.

% We attempted to train the class prototypes along with the backbone but the training quickly collapses.  This is similar to prototypical networks \citep{snell2017prototypical} in which the contrast comes from frozen target embeddings rather than SoftMax layer. Note that our method (CMSF) is non-contrastive as it calculates the target prototypes as the average of nearest neighbors rather than choosing frandom rozen vectors. We change the linear FC layer to be a 2-layer MLP which leads to better representations.

\begin{table}
    \begin{center}
    \caption{\textbf{Linear layer transfer learning evaluation:} Our CMSF model with 200 epochs only outperforms all baselines on transfer learning evaluation. We separate supervised and SSL setting. We copied the results for MoCo v2, MSF, and BYOL-asym from \citet{koohpayegani2021mean}, SimCLR and Xent (1000 epoch) from \citet{chen2020simple}, and BYOL from \citet{grill2020bootstrap}.\\}
    \label{tab:main}
    \scalebox{0.78}{
    \begin{tabular}{l|c|c|c|c|c|c|c|c|c|c|c||c|c}
    \toprule
    Method & \small{Epoch} & Food & \small{CIFAR} & \small{CIFAR} & SUN & Cars & Air- & DTD & Pets & Calt. & Flwr & \textbf{Mean} & \textbf{Linear} \\
    & & 101 & 10 & 100 & 397 & 196 & craft &  &  & 101 & 102 & \textbf{Trans} & \textbf{IN-1k} \\
    \midrule
    Xent & 200 & 67.7 & 89.8 & 72.5 & 57.5 & 43.7 & 39.8 & 67.9 & 91.8 & 91.1 & 88.0 & 71.0 & 77.2 \\
    Xent & 90 & 72.8 & 91.0 & 74.0 & 59.5 & 56.8 & 48.4 & 70.7 & 92.0 & 90.8 & 93.0 & 74.9 & 76.2 \\
    FrzProto & 200 & 71.8 & 92.2 & 75.8 & 60.8 & 67.5 & 58.2 & 72.2 & 91.9 & 93.0 & 94.2 & 77.8 & 75.6 \\
    ProtoNW & 200 & 73.3 & 93.2 & 78.3 & 61.5 & 65.0 & 57.6 & 73.7 & 92.2 & 94.3 & 93.7 & 78.3 & 76.0 \\
    SupCon & 200 & 72.5 & 93.8 & 77.7 & 61.5 & 64.8 & 58.6 & 74.6 & \textbf{92.5} & 93.6 & 94.1 & 78.4 & \textbf{77.5} \\
    Xent & 1000 & 72.3 & 93.6 & 78.3 & 61.9 & 66.7 & 61.0 & \textbf{74.9} & 91.5 & 94.5 & 94.7 & 78.9 & 76.3 \\
    CMSF top-$all$ & 200 & 73.7 & 94.2 & \textbf{78.7} & 62.1 & \textbf{71.7} & \textbf{64.1} & 73.4 & \textbf{92.5} & 94.5 & \textbf{95.8} & \textbf{80.1} & 75.7 \\
    CMSF top-$10$ & 200 & \textbf{74.9} & \textbf{94.4} & \textbf{78.7} & \textbf{62.7} & 70.8 & 63.4 & 73.8 & 92.2 & \textbf{94.9} & 95.6 & \textbf{80.1} & 76.4 \\
    \midrule
    MoCo v2 & 200 & 70.4 & 91.0 & 73.5 & 57.5 & 47.7 & 51.2 & 73.9 & 81.3 & 88.7 & 91.1 & 72.6 & 67.5 \\
    SimCLR & 1000 & 72.8 & 90.5 & 74.4 & 60.6 & 49.3 & 49.8 & \textbf{75.7} & 84.6 & 89.3 & 92.6 & 74.0 & 69.3 \\
    MoCo v2 & 800 & 72.5 & 92.2 & 74.6 & 59.6 & 50.5 & 53.2 & 74.4 & 84.6 & 90.0 & 90.5 & 74.2 & 71.1 \\
    BYOL\small{-asym} & 200 & 70.2 & 91.5 & 74.2 & 59.0 & 54.0 & 52.1 & 73.4 & 86.2 & 90.4 & 92.1 & 74.3 & 69.3 \\
    % MSF \small{-w/s} & 200 & 71.2 & 92.6 & 76.3 & 59.2 & 55.6 & 53.7 & 73.2 & 88.7 & 92.7 & 92.0 & 75.5 & 72.4 \\
    % MSF & 200 & 70.7 & 92.0 & 76.1 & 59.0 & 60.9 & 53.5 & 72.1 & 89.2 & 92.1 & 92.4 & 75.8 & 71.4 \\
    MSF & 200 & 72.3 & \textbf{92.7} & 76.3 & 60.2 & 59.4 & 56.3 & 71.7 & 89.8 & 90.9 & 93.7 & 76.3 & 72.1 \\
    BYOL & 1000  & \textbf{75.3} & 91.3 & \textbf{78.4} & \textbf{62.2} & \textbf{67.8} & \textbf{60.6} & 75.5 & \textbf{90.4} & \textbf{94.2} & \textbf{96.1} & \textbf{79.2} & \textbf{74.3} \\
    \bottomrule
    \end{tabular}
    }
    \end{center}
\end{table}

\subsubsection{Implementation details}
\label{sec:supervised_impl_details}

We experiment with following augmentations: the augmentation from MoCo-v2 \citep{chen2020mocov2} (strong aug), the standard augmentation (std aug) used for supervised Xent ImageNet-1k training \citep{official_pytorch_models} and weak/strong augmentation from MSF \citep{koohpayegani2021mean}. By default both CMSF and SupCon use weak/strong augmentation. All models are trained on supervised ImageNet-1k (IN-1k) for 200 epochs. ResNet-50 \citep{he2016deep} is used as the backbone in all experiments. All models are trained with SGD optimizer (lr=0.05, batch size=256, momentum=0.9, and weight decay=1e-4). Unless mentioned, the learning rate scheduler is cosine. The value of momentum for the moving average key encoder in $0.99$ for CMSF and $0.999$ for SupCon. The MLP architecture for CMSF is a sequence of following layers: linear (2048x4096), batch norm, ReLU, and linear (4096x512). The default memory bank size is 128k for CMSF, SupCon, and ProtoNW. For CMSF, top-$k=10$ is the default. For more details see the appendix. Our main CMSF experiment with 200 epochs takes almost 6 days on four NVIDIA-2080TI GPUs.

\subsubsection{Evaluation}
\label{sec:supervised_evaluation}

We evaluate the supervised pre-trained models by treating them as frozen feature extractors and training a single linear layer on top of them for below listed datasets. 

\textbf{Datasets:} Unlike Xent, methods like SupCon, ProtoNW, and CMSF do not train a linear classifier during the pre-training stage thus we use the pre-training dataset ImageNet-1k (IN-1k) for evaluating the frozen features. The transfer performance is evaluated on the following datasets: Food101 \citep{food101}, SUN397 \citep{sun397}, CIFAR10 \citep{cifar}, CIFAR100 \citep{cifar}, Cars196 \citep{carsdataset}, Aircraft \citep{aircraft}, Flowers (Flwrs102) \citep{flowers}, Pets \citep{pets}, Caltech-101 (Calt101) \citep{caltech101}, and DTD \citep{dtd}. More details about the datasets like train/val/test split sizes can be found in the supplementary material.

\textbf{Linear IN-1k:} We follow the linear evaluation setup from CompRess \citep{abbasi2020compress}. The features are normalized to have unit $\ell_2$ norm and then scaled and shifted to have zero mean and unit variance for each dimension. We use SGD optimizer (lr=0.01, epochs=40, batch size=256, weight decay=1e-4, and momentum=0.9). Learning rate is multiplied by 0.1 at epochs 15 and 30. We use standard supervised ImageNet augmentations \citep{official_pytorch_imagenet_train} during training. For Xent and FrzProto, we use the linear classifier trained during pre-training.

\textbf{Transfer:} We follow the procedure outlined in BYOL \citep{grill2020bootstrap} and SimCLR \citep{chen2020simple} to train a single linear layer on top of the frozen backbone. We tune the hyperparameters for each dataset independently based on the validation set accuracy, and report the final accuracy on the held-out test set. More details about the training procedure can be found in the Appendix.

\subsubsection{Results}

The results for the IN-1k dataset pre-training are reported in Table \ref{tab:main}. First, we find that CMSF has the best transfer evaluation results. Second, as shown by SupCon and Xent (200 epoch version), improvements on the Linear ImageNet-1k evaluation do not always translate to transfer evaluation. Third, methods inspired from SSL like FrzProto, SupCon, and CMSF are generally better than their Xent counterparts for similar number of epochs. Fourth, CMSF is highly competitive with a self-supervised method like BYOL which is trained for 5 times more epochs (200 vs 1000 epochs). Fifth, it is surprising that the FrzProto baseline can achieve such a high performance. Finally, our method performs better at fine-grained datasets, e.g., Cars196 and Aircraft, which we believe is due to preserving multi-modal distribution of the categories.

\begin{table}[t]
    \begin{center}
    \caption{{\bf Ablations of baselines and CMSF:} All experiments use 200 epochs if not mentioned and use ImageNet-1k dataset. \textbf{(a)} More epochs does not improve transfer accuracy for Xent. Thus, the model available from PyTorch \citep{official_pytorch_models} (last row) has the best transfer accuracy; \textbf{(b)} We add components of our method to improve SupCon baseline. The baseline implementation of SupCon uses std. aug and 16k memory size and it does not include the target embedding $u$ in the positive set. \textbf{(c)} Using a MLP head improves FrzProto a lot as it allows the post-MLP features to adapt to the regression task, but allows pre-MLP features to be generalizable; \textbf{(d)} We find that our method is not very sensitive to the size of memory bank or in top-$k$; \textbf{(e)} Interestingly, excluding the target embedding $u$ from $\hat M$ does not hurts the results. Note that when we do not include the target, the nearest neighbors are still chosen based on the distance to the target, so they will be close to the target. \textbf{(f)} We report the results of our method by with varying the amount of labeled data on ImageNet-1k. We find that only 50\% of labeled data is sufficient to reach on-par performance of the fully supervied model. The first row is equivalent to self-supervised MSF, so the numbers are copied from \citet{koohpayegani2021mean}}
    \label{tab:all_ablations}
    \vspace{.1in}
    \scalebox{0.93}
    {
    \begin{tabular}{lcc}
        \toprule
        Method & Mean & Linear \\
        & Trans & IN-1k \\
        
        \midrule
        % \multicolumn{3}{c}{Xent} \\
        % \midrule
        \textit{(a) Xent} & & \\
        lr=$0.05$, cos, epochs=$200$, strong aug. & 71.5 & 77.2 \\
        lr=$0.05$, cos, epochs=$200$, std. aug. & 71.0 & 77.3 \\
        lr=$0.10$, cos, epochs=$200$, strong aug. & 72.3 & 77.1 \\
        lr=$0.05$, cos, epochs=$90$, std. aug. & 72.4 & 76.8 \\
        lr=$0.10$, cos, epochs=$90$, std. aug. & 74.0 & 76.7 \\
        lr=$0.10$, step, epochs=$90$, std. aug. & 74.9 & 76.2 \\
        
        \midrule
        % \multicolumn{3}{c}{SupCon} \\
        % \midrule
        \textit{(b) SupCon} & & \\
        Base SupCon & 77.2 & 77.9 \\
        + change to strong aug. & 77.9 & 77.4 \\
        + add target to positive set & 77.8 & 77.4 \\
        + change to weak/strong aug. & 77.8 & 77.2 \\
        + increase mem size to 128k & 78.4 & 77.5 \\
        
        \midrule
        % \multicolumn{3}{c}{FrzProto} \\
        % \midrule
        \textit{(c) FrzProto} & & \\
        FC = Linear & 43.3 & 74.0 \\
        FC = MLP & 77.8 & 75.6 \\

        \bottomrule
    \end{tabular}
    }
    \quad
    \scalebox{0.9}
    {
    \begin{tabular}{lcc}
        \toprule
        Method & Mean & Linear \\
        & Trans & IN-1k \\

        \midrule
        % \multicolumn{3}{c}{CMSF} \\
        % \midrule
        \textit{(d) CMSF} & & \\
        top-$1$ \small{(BYOL-asym)}  & 74.3 & 69.3 \\
        mem=128k, top-$2$ & 78.4 & 76.2 \\
        mem=128k, top-$10$ & 80.1 & 76.4 \\
        mem=128k, top-$20$ & 79.9 & 76.3 \\
        mem=128k, top-$all$ & 80.1 & 75.7 \\
        mem=512k, top-$10$ & 79.9 & 76.2 \\
        mem=512k, top-$20$ & 80.1 & 76.3 \\
        
        % \midrule
        % % \multicolumn{3}{c}{CMSF} \\
        \midrule
        \textit{(e) CMSF} & & \\
        target in top-$10$ & 80.1 & 76.4 \\
        target not in top-$10$ & 80.3 & 76.4 \\
        
        % \midrule
        % % \multicolumn{3}{c}{CMSF} \\
        \midrule
        \textit{(f) CMSF} & & \\
        labels 0\% \small{(MSF)} & 75.5 & 72.4 \\
        labels 10\% & 77.8 & 73.0 \\
        labels 20\% & 78.3 & 73.8 \\
        labels 50\% & 79.4 & 75.3 \\
        labels 100\% & 80.1 & 76.4 \\

        \bottomrule
        
    \end{tabular}
    }
    \end{center}
\end{table}

\subsubsection{Ablations}
\label{sec:ablations}

We explore different design choices and parameters of our method and baselines. We add the techniques used for our methods to the baselines to isolate the effect of different losses. The results are reported in Table \ref{tab:all_ablations}. Training and evaluation details are the same as in Sections \ref{sec:supervised_impl_details} and \ref{sec:supervised_evaluation}.

While traditional semi-supervised learning \citep{tarvainen2017mean} focuses on only improving the performance on the training dataset, we explore how the amount of labeled data in the dataset influences generalization of the representations to other datasets. Since we use the constraint to limit the search space only, our method can easily benefit from the constraint even if it is available only for a subset of the data. Hence, the method can be easily extended to semi-supervised setting by simply setting $\hat M=M$ for the unlabeled data. We use two equal-size, separate memory banks for labeled and unlabeled data so that one cannot dominate the whole memory bank. The results are reported in section (f) of  Table \ref{tab:all_ablations}.

\subsection{Noisy Supervised Setting}
\label{sec:noisy_section}
Our method can handle noisy labels since it considers only top NN results as shown in Figure \ref{fig:visualize_nns}. To add noise to the dataset, the labels for a certain percentage of images are corrupted randomly and kept constant throughout training. We consider, 5\%, 10\%, 25\% and 50\% label corruption (noise) rates. The corrupted labels are provided in the code (supplementary). For faster experiments, we use the supervised ImageNet-100 (IN-100) split \citep{tian2019cmc}. Our main goal is to show that top-$k$ is more robust than top-$all$. %We also compare with Xent in order to study the effect of contrast.

\textbf{Training and evaluation details:} The implementation and evaluation details are the same as in Sections \ref{sec:supervised_impl_details} and \ref{sec:supervised_evaluation} except that we use 500 epochs following LooC \citep{xiao2020contrastive}, and use linear evaluation on clean IN-100 for both our method and the baseline. We use a memory bank of size 64k. We observed that the transfer accuracy of Xent on 25\% and 50\% degrades with longer training, so we report Xent results with 100 epochs only for the corrupted data.

\begin{figure*}[t]
\centering
\includegraphics[width=0.9\linewidth]{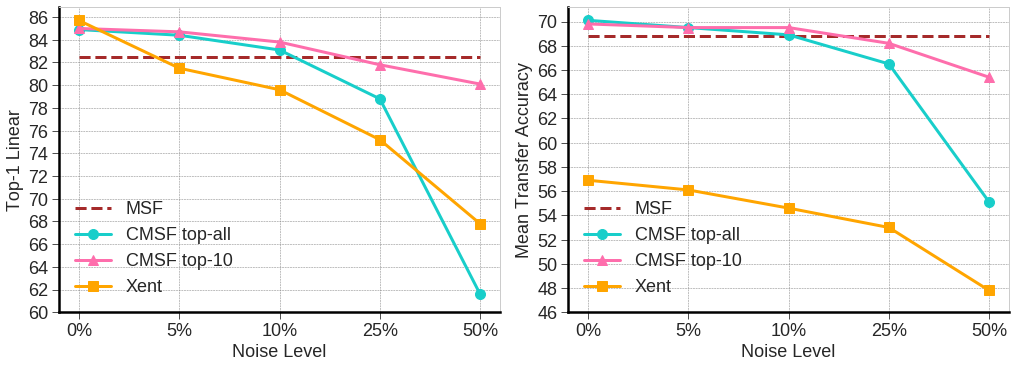}
\caption{\textbf{Noisy supervised setting on ImageNet-100:} Our method is more robust to noisy annotation compared to Xent. Also, using top-$all$ results in degradation since all images from a single category are not guaranteed to be semantically related. Mean Transfer Accuracy is average accuracy of each model over 10 transfer dataset in our settings.}
\label{fig:main_noisy}
\vspace{-.1in}
\end{figure*}

\textbf{Results:} 
The results are reported in the Figure \ref{fig:main_noisy}. We observe that by increasing the amount of noise, the accuracy of our method drops less compared to the baseline. The gap is larger for the transfer learning evaluation. Moreover, CMSF top-$all$ degrades more compared to top-$10$. This shows that in the presence of a noisy constraint, it is better to only pull locally close embeddings close together. Since all embeddings of the same class are not guaranteed to contain the same semantic content, doing top-$all$ pulls semantically unrelated embeddings close together which hurts the quality of representations. This is shown in Figure \ref{fig:visualize_nns}.

% On linear IN-100 evaluation, our Xent has 85.7\% accuracy while LooC \citep{xiao2020contrastive} has 83.7\%.

\subsection{Cross-modal Constraint}

In the previous Section \ref{sec:noisy_section} we showed that our method (CMSF with top-$k$) is robust when the constraint is noisy. Here, we explore another such noisy constraint: an already trained SSL model in another modality. Following \citet{han2021selfsupervised}, we use split-1 of UCF-101 \citep{soomro2012ucf101} (13k videos) as the unlabeled dataset. 
We use similar augmentation and pre-processing as \citet{han2021selfsupervised} and calculate optical-flow using unsupervised TV-L1 \citep{Zach2007ADB} algorithm. We first train two SSL models on RGB and Flow modalities separately using InfoNCE method \citep{oord2018representation,han2021selfsupervised}. Then we continue training on one modality while freezing the other modality and using it as a constraint. In training the flow network using RGB network as constraint, we sample $n$ nearest neighbors in RGB's memory bank and then we search for top-$k$ nearest neighbors among those samples in Flow's memory bank. We use the code from \citet{han2021selfsupervised} for linear evaluation. We report top-1 accuracy for linear classification and recall@1 for retrieval on the extracted features of frozen networks.

\textbf{Results:}
We show the results in Table \ref{tab:co_training}. All experiments use spatio-temporal 3D data either in RGB or flow format. Our method outperforms all the baselines including CoCLR \citep{han2021selfsupervised} in Flow and is 3.6 points less accurate that CoCLR on RGB modality.

\begin{table}[ht!]
    \centering
    \caption{{\bf Cross-modal constraint: } We continue training each SSL model for 200 epochs using MSF \citep{koohpayegani2021mean} for a fair comparison. ``Init" column shows what model has been used to initialize the training while ``Conatraint'' column shows what model is used to provide the constraint. Note that CoCLR \citep{han2021selfsupervised} also uses another modality as a constrain in the form of contrastive learning. Rows [1-4] are copied from \citet{han2021selfsupervised}.\\}
    \label{tab:co_training}
    \scalebox{0.75}{
    \begin{tabular}{llcccccc}
        \toprule
        Index & Model  & Modality & Init & Constraint & Epochs & R@1 & Linear \\
        \midrule
        1 & Sup Xent  & RGB & - & - & - & 73.5 & 77.0 \\
        2 & Sup UberNCE   & RGB & - & - & - & 71.6 & 78.0 \\
        \midrule
        3 & CoCLR\tiny{k=5}   & RGB & - & Flow & 500 & 51.8 & 70.2 \\
        4 & CoCLR \tiny{k=5}   & Flow & - & RGB & 500 & 48.4 & 67.8 \\
        \midrule
        5 & InfoNCE  & RGB & - & - & 400 & 35.5 & 47.9 \\
        6 & MSF  \tiny{k=5}  & RGB & 5 & - & +200 & 39.6 & 50.8 \\
        7 & CMSF \tiny{n=10, k=5} & RGB & 5 & 10 & +100 & 45.8 & 58.1 \\
        %9 & MSF-Const \tiny{N=20, K=5} & RGB & 1 & 2 & +100 & 46.2 & 58.0 \\
        8 & CMSF \tiny{n=10, k=5}  & RGB & 5 & 14 & +100 & 46.2 & 58.1 \\
        %5 & MSF-Const \tiny{N=5, K=5} & RGB & 1 & 2 & +100 & 46.8 & 59.1 \\
        %15 & MSF-Const \tiny{N=5, K=5} & RGB & 1 & 16 & +100 & 48.9 & 59.7 \\
        9 & CoCLR \tiny{k=5} & RGB & 5 & 10 & +100 & \textbf{49.8} & \textbf{61.0} \\
        
        \midrule
        10 & InfoNCE & Flow & - & - & 400 & 45.3 & 66.1 \\
        11 & MSF \tiny{k=5}  & Flow & 10 & - & +200 & 47.3 & 64.7 \\
        12 & CoCLR \tiny{k=5}  & Flow & 10 & 9 & +100 & 50.0 & 67.3 \\
        %6 & MSF-Const \tiny{N=5, K=5} & Flow & 2 &  5 & +100 & 53.2 & 70.1 \\
        %10 & MSF-Const \tiny{N=20, K=5} & Flow & 2 & 9 & +100 & 54.0 & 70.2 \\
        13 & CMSF \tiny{n=10, k=5} & Flow & 10 & 9 & +100 & 54.1 & \textbf{71.2} \\
        %16 & MSF-Const \tiny{N=5, K=5} & Flow & 2 & 1 & +100 & 55.1 & 69.8 \\
        14 & CMSF \tiny{n=10, k=5}  & Flow & 10 & 5 & +100 & \textbf{55.6} & 70.2 \\
        \bottomrule
    \end{tabular}
    }
    \vspace{-.1in}
\end{table}

\textbf{Ablation (effect of $n$):} We show the effect of $n$ in Table \ref{tab:co_training_ablation}. It is interesting that the 6th row is better than the 4th row. We hypothesize increasing $n$ helps when the constraint is not very accurate (RGB in this case) by loosening the constraint. This is aligned with our intuition.

\textbf{Implementation Details.}
For cross-modal experiments, we use S3D \citep{xie2018rethinking} architecture with the input size of 128 pixels. We initialize from the pretrained weights of InfoNCE (400-epoch) released by \citep{han2021selfsupervised}. We use following settings for our method: memory bank of size $8,192$, $n=10$, $k=5$, batch size $128$, weight decay $1e-5$, initial lr of $0.001$, and learning rate decay by factor of $10$ at epoch 80. We train each modality for additional 100 epochs using PyTorch Adam optimizer. For a fair comparison, we run CoCLR using their official code by initializing it from the same model as ours.

\subsection{Constraining with 2D SSL models} 
We show that self-supervised 2D ResNet50 model pretrained on ImageNet-1k can be used as a constraint for 3D video SSL models. We initialize S3D backbone from the self-supervised InfoNCE model and continue training it with ImageNet-1k pretrained, SSL ResNet50 as the constraint. We randomly select one frame of input video and feed it to the ResNet50 backbone to get the features for the constraint. We use $n=10$ and $k=5$. We also evaluate the 2D backbone on the center frame of the video only. The results are shown in Table \ref{tab:2dimagetoflow}. 
Interestingly, our CMSF Flow model outperforms the 2D model by $3.2$ points in R@1 and $1.7$ point in Linear evaluation. Again since $n > k$, the NN search on the 3D model can search for the best samples rather than fully relying on the constraint. 

\textbf{Implementation Details.} We use the same details as the cross-modal experiments except that the initial lr is $0.002$ with decaying by a factor of $10$ at epoch 180.

\begin{table}
\parbox{.50\linewidth}{
    \centering
    \caption{{\bf Ablation for the effect of n for cross-modal setting: } By comparing Rows 4 and 6, we can see that using $n>k$ helps the model when the constraint is less accurate (more noisy).\\}
    \label{tab:co_training_ablation}
    \scalebox{0.68}{
    \begin{tabular}{llcccccc}
        \toprule
        Idx & Model & Mod- & Init & Cons- & Epochs & R@1 & Linear \\
        & & ality & & traint & & & \\
        \midrule
        1 & InfoNCE & RGB & - & - & 400 & 35.5 & 47.9 \\
        2 & InfoNCE & Flow & - & - & 400 & 45.3 & 66.1 \\
        
        3 & CMSF \tiny{n=5, k=5} & RGB & 1 & 2 & +100 & 46.8 & 59.1 \\
        4 & CMSF \tiny{n=5, k=5} & Flow & 2 &  3 & +100 & 53.2 & 70.1 \\
       
        5 & CMSF \tiny{n=10, k=5} & RGB & 1 & 2 & +100 & 45.8 & 58.1 \\
        6 & CMSF \tiny{n=10, k=5} & Flow & 2 & 5 & +100 & \textbf{54.1} & \textbf{71.2} \\
        
        7 & CMSF \tiny{n=20, k=5} & RGB & 1 & 2 & +100 & 46.2 & 58.0 \\
        8 & CMSF \tiny{n=20, k=5} & Flow & 2 & 7 & +100 & 54.0 & 70.2 \\
        \bottomrule
    \end{tabular}
    \vspace{-.1in}
    }
}
\hfill
\parbox{.46\linewidth}{
    \centering
    \caption{{\bf Constraining with 2D SSL models:} Our method can benefit from constraints that come from an SSL 2D model trained on ImageNet-1k. ``CF'' refers to running the 2D model on the center frame of the video.\\}
    \label{tab:2dimagetoflow}
    \scalebox{0.68}{
    \begin{tabular}{llclcccccc}
        \toprule
        Idx & Model & Arch & Mod- & Cons- & R@1 & Linear \\
        & & & ality & traint & & \\
        \midrule
        %1 & 2D-XE & R50 &  RGB \tiny{CF} & -  & 55.9 & 69.6 \\
        %2 & 2D-CMSF &  R50 &  RGB \tiny{CF} & -  & \textbf{62.2} & \textbf{77.1} \\
        %\midrule
        1 & 2D-MSF &  R50 &  RGB \tiny{CF} & - & 57.2 & 71.5 \\
        \midrule
        2 & CMSF & S3D & RGB & 1 & 48.9 & 59.1 \\
        3 & CMSF & S3D & Flow & 1 & \textbf{60.4} & \textbf{73.2} \\
        \bottomrule
    \end{tabular}
    }
}
\vspace{-.1in}
\end{table}

\section{Related Work}
{\bf Supervised learning:} Cross-entropy is a well known loss function for supervised learning \citep{NIPS1987_eccbc87e, levin1988accelerated, rumelhart1986learning}. %n cross-entropy, each class has one-hot probability distribution. 
Cross-entropy is contrastive in nature since ground truth probability of each class is either $0$ or $1$. One drawback of Cross-entropy is its lack of robustness to noisy labels \citep{zhang2018generalized,sukhbaatar2015training}. %Further, some works have tried to mitigate drawbacks of cross-entropy. For example,
\citet{szegedy2015rethinking,muller2020does} address the issue of hard labeling (one-hot labels) with label smoothing, \citet{hinton2015distilling,bagherinezhad2018label,furlanello2018born} replace hard labels with prediction of pretrained teacher, and \citet{zhang2018mixup,yun2019cutmix} propose an augmentation strategy to train on combination of instances and their labels. Another line of works \citep{goldberger2004neighbourhood,salakhutdinov2007learning} have attempted to learn representations with good kNN performance. Supervised Contrastive Learning (SupCon) \citep{khosla2020supervised} and \citet{wu2018improving} improve upon \citet{goldberger2004neighbourhood} by changing the distance to inner product on $\ell_2$ normalized embeddings. Our method is different as it does not use negative samples which makes it non-contrastive. Moreover, we focus on learning transferable representations instead of just focusing on the pretraining task. We also show that our method is robust to noisy supervision.

{\bf Constrained clustering:} Constrained clustering has been studied before \citet{basu2008constrained,legendre1987constrained,ganccarski2020constrained}. \citet{zhang2019framework} adds various constraints to deep k-means-like clustering. \citet{jia2020constrained} uses constraints with Graph-Laplacian PCA. \citet{anand2013semi} generalizes mean-shift algorithm with kernel learning and pairwise constraints. Our method is a simple non-contrastive mean-shift algorithm that is inspired by the recent success in self supervised learning literature by comparing different augmentations of the same image.

{\bf Metric learning:} The goal of metric learning is to train a representation that puts two instances close in the embedding space if they are semantically close. Two important methods in metric learning are: triplet loss \citep{chopra2005learning,weinberger2006distance,schroff2015facenet} and contrastive loss \citep{sohn2016improved,bromley1993signature}. Both use positive and negative samples, but the number of negatives is larger in contrastive losses. Because of the negative samples, these losses are contrastive in nature. It has been shown that metric learning methods perform well on tasks like image retrieval \citep{Wu_2017_ICCV} and few-shot learning \citep{vinyals2017matching,snell2017prototypical}. One of these few-shot learning works is prototypical networks \citep{snell2017prototypical} which is similar to a contrastive version of our method with top-$all$.

{\bf Self-supervised learning (SSL): } Here, we want to learn representations without any annotations. One way to learn form unlabeled data is by solving a pretext task. Examples of a pretext task are colorization \citep{zhang2016colorful}, jigsaw puzzle \citep{noroozi2016unsupervised}, counting \citep{noroozi2017representation} and rotation prediction \citep{gidaris2018unsupervised}. Another class of SSL methods are based on instance discrimination \citep{dosovitskiy2014discriminative}. The idea is to classify each image as its own class. Some methods adopted the idea of contrastive learning for instance discrimination \citep{he2020momentum,chen2020simple,caron2018deep,caron2021unsupervised}. BYOL \citep{grill2020bootstrap} proposes a non-contrastive approach by removing the negative set from contrastive SSL methods and simply regressing target view of an image from the query view. MSF \citep{koohpayegani2021mean} generalizes BYOL by regressing target view and its NNs. Our idea is adopted from MSF \citep{koohpayegani2021mean} by using an additional source of knowledge to constrain the NN search space for the target view.

\textbf{Multi-modal self-supervised learning.} Similar to self-supervised learning on single modality, the goal is to learn a rich representation with more than one modality per instance. One approach is to use corresponding (audio, frame) pairs \citep{alwassel2020selfsupervised,arandjelovic2017look,arandjelovic2018objects, korbar2018cooperative, patrick2020multimodal,piergiovanni2020evolving} with contrastive loss. Another approach is to use video and text narration \citep{miech2020endtoend}.

We adopt the idea of co-training from \citep{blum2000CoTraining} and CoCLR \citep{han2021selfsupervised} where two networks help in training each other. In our case, we train two SSL models in Flow and RGB modalities. We use knowledge of one modality as a constrain to train the other modality.

% In the original mean shift clustering algorithm \citep{cheng1995mean,comaniciu2002mean}, each point is pulled towards its neighbors for the purpose of clustering. MSF uses a form of this grouping based on kNNs for self-supervised learning. While, our method limits the search of kNNs based on a hard constraint that is coming from either labels or another modality. The core of our method is "shift"ing an embedding towards the "mean" of its constrained nearest neighbors. \hamed{feel free to remove this}

\section{Conclusion}
Adopting ideas from SSL literature, we introduce a non-contrastive generalized framework that learns rich representations by grouping similar images together while taking advantage of some other source of knowledge if available, e.g., labels or another modality. We show that our method outperforms the baselines on various settings including supervised with clean or noisy labels, and video SSL settings.

% \soroush{remove this part?}One limitation of this submission is that we do not study why the model does not collapse to a single embedding. The same phenomenon is observed on BYOL and MSF, so needs further study.

\textbf{Ethics Statement:}
We are introducing a method for a core problem in computer vision: representation learning. Hence, similar to most other AI algorithms, our method can be exploited by adversaries for unethical applications. Since we are not focusing on a particular application, we cannot list any specific ethical issues. On the positive note, since our method can achieve similar results with fewer or noisy labeled data, it may facilitate making AI more accessible to a large community leading to democratizing AI.

%Our Most AI algorithms can be exploited by adversaries for unethical applications. Our method is not an exception even though we are introducing a solution for a core problem in representation learning that is not specifically tuned for a particular unethical application. 

\textbf{Reproducibility Statement:}
To make our work reproducible, we report details of the implementation for each section. For Supervised section, details of implementation are in Section 2.1.2. Details of transfer learning benchmark are in Appendix Section A.3. More implementation details of each baseline are in Appendix Section A.1.  Moreover, we submit our code as supplementary material.

{\bf Acknowledgment:} 
This material is based upon work partially supported by the United States Air Force under Contract No. FA8750‐19‐C‐0098, funding from SAP SE, and also NSF grant numbers 1845216 and 1920079. Any opinions, findings, and conclusions or recommendations expressed in this material are those of the authors and do not necessarily reflect the views of the United States Air Force, DARPA, or other funding agencies.

\bibliography{iclr2022_conference}
\bibliographystyle{iclr2022_conference}

\appendix
\section{Appendix}
% \section*{Appendix}
\renewcommand{\thefigure}{A\arabic{figure}}
\renewcommand{\thetable}{A\arabic{table}}
\setcounter{figure}{0}
\setcounter{table}{0}

\subsection{Implementation Details of Baselines (Section 2.1.1)}
\medskip

The MLP architecture for FrzProto is: linear (2048x2048), batch norm, ReLU, and linear (2048x2048). For SupCon baseline, it is: linear (2048x2048), batch norm, ReLU, and linear (2048x128). For FrzProto, the class prototype embeddings are a matrix of size 1000x2048 for ImageNet-1k. For optimizing SupCon baseline, following [31], we use the first 10 epochs for learning-rate warmup. For both SupCon and ProtoNW, the temperature is $0.1$.

\subsection{Semi-Supervised Setting (Section 2.1.5)}
\medskip
In training, we have two memory banks of size 128K each. All samples of a mini-batch are pushed to the unlabeled memory bank while only labeled samples are pushed to the labeled memory bank. Unconstrained loss (Self-Supervised MSF) or constrained loss (Supervised) is calculated for each sample depending on the existence of its label, and the final loss is the mean of losses for the whole mini-batch. All other implementation details are the same as in Section 2.1.2. We report the extended version of Table 2(f) for all 10 transfer dataset. Results are in Table \ref{tab:semi_supervised_extended}. 

\begin{table}[h!]
    \begin{center}
    \caption{\textbf{Semi-supervised representation learning with CMSF.} We report the results of our method by with varying the amount of labeled data on ImageNet-1k. We find that only 50\% of labeled data is sufficient to reach on-par performance of the fully supervied model. The first row is equivalent to self-supervised MSF, so the numbers are copied from \cite{koohpayegani2021mean}}
    \vspace{.15in}
    %*: is equivalent to self-supervised MSF and the numbers are copied from \cite{koohpayegani2021mean}.\\
    \label{tab:semi_supervised_extended}
    \scalebox{0.85}{
    \begin{tabular}{l|c|c|c|c|c|c|c|c|c|c||c|c}
    \toprule
    Labeled & Food & CIFAR & CIFAR & SUN & Cars & Air- & DTD & Pets & Calt. & Flwr & \textbf{Mean} & \textbf{Linear} \\
    Split & 101 & 10 & 100 & 397 & 196 & craft &  &  & 101 & 102 & \textbf{Trans} & \\
    \midrule
    0\% & 71.2 & 92.6 & 76.3 & 59.2 & 55.6 & 53.7 & 73.2 & 88.7 & 92.7 & 92.0 & 75.5 & 72.4 \\ 
    10\% & 71.6 & 93.6 & 78.1 & 61.0 & 62.0 & 59.2 & 73.4 & 91.5 & 93.1 & 94.4 & 77.8 & 73.0 \\
    20\% & 73.3 & 93.2 & 77.8 & 61.3 & 64.5 & 60.1 & 73.6 & 91.0 & 93.4 & 95.0 & 78.3 & 73.8 \\
    50\% & 74.1 & 93.8 & 79.4 & 62.1 & 68.6 & 63.1 & 73.0 & 91.6 & 93.5 & 95.0 & 79.4 & 75.3 \\
    100\% & 74.9 & 94.4 & 78.7 & 62.7 & 70.8 & 63.4 & 73.8 & 92.2 & 94.9 & 95.6 & 80.1 & 76.4 \\
    \bottomrule
    \end{tabular}
    }
    \end{center}

\end{table}

\subsection{Implementation Details of Transfer Evaluation (Section 2.1.3)}
We use the LBFGS optimizer (max\_iter=20, and history\_size=10) along with the Optuna library \citep{akiba2019optuna} in the Ray hyperparameter tuning framework \citep{liaw2018tune}. Each dataset gets a budget of 200 trials to pick the best parameters on validation set. The final accuracy is reported on a held-out test set by training the model on the train+val split using the best hyperparameters. The hyperparameters and their search spaces (in loguniform) are as follows: iterations $\in [0, 10^3]$, lr $\in [10^{-6}, 1]$, and weight decay $\in [10^{-9}, 1]$. We also show that we can reproduce the transfer results for BYOL \citep{grill2020bootstrap} and SimCLR \citep{chen2020simple} with our framework. The features are extracted with the following pre-processing for all datasets: resize shorter side to 256, take a center crop of size 224, and normalize with ImageNet statistics. No training time augmentation was used.

\begin{table}[ht!]
    \begin{center}
    \caption{\textbf{Transfer dataset details:} Train, val, and test splits of the transfer datasets are listed in this table. \textbf{Test split: } We follow the details in [32]. For Aircraft, DTD, and Flowers datasets, we use the provided test sets. For Sun397, Cars, CIFAR-10, CIFAR-100, Food101, and Pets datasets, we use the provided val set as the hold-out test set. For Caltech-101, 30 random images per category are used as the hold-out test set. \textbf{Val split: } For DTD and Flowers, we use the provided val sets. For other datasets, the val set is randomly sampled from the train set. For transfer setup, to be close to BYOL [25], the following val set splitting strategies have been used for each dataset: Aircraft: 20\% samples per class. Caltech-101: 5 samples per class. Cars: 20\% samples per class. CIFAR-100: 50 samples per class. CIFAR-10: 50 samples per class. Food101: 75 samples per class. Pets: 20 samples per class. Sun397: 10 samples per class.\\}
    \label{tab:appendix_transfer_dset_details}
    \scalebox{0.77}{
    \begin{tabular}{lrrrrrrr}
        \toprule
        Dataset & Classes & Train samples & Val samples & Test samples & Accuracy measure & Test provided \\
        \midrule
        Food101 [11] & 101 & 68175 & 7575 & 25250 & Top-1 accuracy & - \\
        CIFAR-10 [35] & 10 & 49500 & 500 & 10000 & Top-1 accuracy & - \\
        CIFAR-100 [35] & 100 & 45000 & 5000 & 10000 & Top-1 accuracy & -\\
        Sun397 (split 1) [65] & 397 & 15880 & 3970 & 19850 & Top-1 accuracy & - \\
        Cars [34] & 196 & 6509 & 1635 & 8041 & Top-1 accuracy & - \\
        Aircraft [39] & 100 & 5367 & 1300 & 3333 & Mean per-class accuracy & Yes \\
        DTD (split 1) [18] & 47 & 1880 & 1880 & 1880 & Top-1 accuracy & Yes \\
        Pets [47] & 37 & 2940 & 740 & 3669 & Mean per-class accuracy & - \\
        Caltech-101 [20] & 101 & 2550 & 510 & 6084 & Mean per-class accuracy & - \\
        Flowers [42] & 102 & 1020 & 1020 & 6149 & Mean per-class accuracy & Yes \\
        \bottomrule
    \end{tabular}
    }
    \end{center}
\end{table}

\subsection{Detailed results for noisy supervised setting (Section 2.2)}

\begin{table}[th!]
    \begin{center}
    \caption{\textbf{Noisy supervised setting on ImageNet-100:} Our method is more robust to noisy annotation compared to Xent. Also, using top-$all$ results in degradation since all images from a single category are not guaranteed to be semantically related.\\}
    \label{tab:main_noisy}
    \scalebox{0.77}{
    \begin{tabular}{l|r|c|c|c|c|c|c|c|c|c|c||c|c}
    \toprule
    Method & Noise & Food & CIFAR & CIFAR & SUN & Cars & Air- & DTD & Pets & Calt. & Flwr & \textbf{Mean} & \textbf{Linear} \\
    & & 101 & 10 & 100 & 397 & 196 & craft &  &  & 101 & 102 & \textbf{Trans} & \textbf{IN-100} \\
    \midrule
    Xent & 0\% & 53.6 & 81.9 & 61.1 & 37.8 & 25.7 & 29.5 & 56.9 & 69.7 & 70.2 & 82.3 & 56.9 & 85.7 \\
    CMSF \small{top-$all$} & 0\% & 61.6 & 88.2 & 68.5 & 49.9 & 54.6 & 52.7 & 64.7 & 82.2 & 89.6 & 89.1 & 70.1 & 84.9 \\
    CMSF \small{top-$10$} & 0\% & 62.6 & 86.8 & 66.2 & 50.5 & 54.7 & 51.0 & 64.6 & 82.4 & 88.5 & 90.4 & 69.8 & 85.0 \\
    \midrule
    Xent & 5\% & 46.5 & 81.1 & 58.1 & 35.8 & 27.5 & 36.0 & 58.7 & 67.5 & 73.3 & 77.0 & 56.1 & 81.5 \\
    CMSF \small{top-$all$} & 5\% &  60.3 & 87.5 & 66.4 & 49.1 & 55.5 & 53.0 & 64.8 & 80.9 & 87.3 & 89.9 & 69.5 & 84.4 \\
    CMSF \small{top-$10$} & 5\% & 61.6 & 86.8 & 67.4 & 49.6 & 55.8 & 51.2 & 63.4 & 81.5 & 86.7 & 90.6 & 69.5 & 84.7 \\
    \midrule
    Xent & 10\% & 44.1 & 79.5 & 56.1 & 32.4 & 26.1 & 34.5 & 56.1 & 69.7 & 72.5 & 75.1 & 54.6 & 79.6 \\
    CMSF \small{top-$all$} & 10\% & 59.4 & 86.4 & 66.0 & 48.8 & 55.0 & 51.4 & 64.7 & 80.1 & 87.8 & 89.0 & 68.9 & 83.1 \\
    CMSF \small{top-$10$} & 10\% & 60.9 & 87.2 & 66.9 & 49.4 & 54.2 & 51.4 & 65.5 & 80.6 & 88.5 & 90.0 & 69.5 & 83.8 \\
    \midrule
    Xent & 25\% & 49.0 & 77.2 & 54.5 & 30.6 & 25.9 & 30.7 & 53.1 & 66.6 & 64.1 & 77.8 & 53.0 & 75.2\\
    CMSF \small{top-$all$} & 25\% & 56.4 & 85.7 & 64.2 & 46.0 & 53.6 & 49.6 & 62.7 & 74.2 & 85.2 & 87.4 & 66.5 & 78.8 \\
    CMSF \small{top-$10$} & 25\% & 58.9 & 85.2 & 64.9 & 47.8 & 55.0 & 50.6 & 64.0 & 80.0 & 86.3 & 89.7 & 68.2 & 81.8 \\
    \midrule
    Xent & 50\% & 44.4 & 72.3 & 51.3 & 31.1 & 21.4 & 24.9 & 46.0 & 57.4 & 56.0 & 73.0 & 47.8 & 67.8 \\
    CMSF \small{top-$all$} & 50\% & 44.7 & 79.3 & 54.9 & 35.2 & 35.7 & 41.2 & 54.9 & 54.6 & 75.3 & 75.1 & 55.1 & 61.6 \\
    CMSF \small{top-$10$} & 50\% & 58.7 & 85.7 & 64.2 & 47.5 & 51.6 & 50.5 & 62.0 & 77.3 & 86.8 & 70.1 & 65.4 & 80.1 \\
    \bottomrule
    \end{tabular}
    }
    \end{center}
\end{table}

\end{document}